\documentclass{article}

\usepackage{arxiv}

\usepackage{url}            
\usepackage{booktabs}       
\usepackage{amsfonts}       
\usepackage{nicefrac}       
\usepackage{microtype}      

\def\bng{\bngx}

\font\bngx=bang10

\def\*#1*#2{o\null{#2}{#1}}

\def\sh#1{\setbox0=\hbox{#1}%
     \kern-.02em\copy0\kern-\wd0
     \kern.04em\copy0\kern-\wd0
     \kern-.02em\raise.0433em\box0 }

 \usepackage{hyperref}
 \hypersetup{colorlinks=true,linkcolor=citeblue,citecolor=citeblue,urlcolor=citeblue}

\usepackage{cite}
\usepackage{amsmath,amssymb,amsfonts}
\usepackage{cleveref}
\usepackage{subcaption}
\usepackage{makecell}

\usepackage[T1]{fontenc}
\usepackage{lmodern}

\usepackage{lipsum}
\usepackage{xcolor}
\usepackage{graphicx}
\usepackage{geometry}
\usepackage{array}
\usepackage{multirow}
\usepackage{xcolor}
\definecolor{citeblue}{RGB}{0,0,255}

\graphicspath{ {./images/} }

\title{Assessing the Level of Toxicity Against Distinct Groups in Bangla Social Media Comments: A Comprehensive Investigation}

\author{
\\
\newline
\\
Mukaffi Bin Moin\textsuperscript{*},
Pronay Debnath,
Usafa Akther Rifa,
Rijeet Bin Anis
\\
\\Ahsanullah University of Science and Technology, Dhaka, Bangladesh.
\\
\\
\bigskip
*Corresponding author(s). E-mail(s): \texttt{\textcolor{blue}{mukaffi28@gmail.com}}\\
Contributing authors: \texttt{\textcolor{blue}{pronaydebnath99@gmail.com}}; \texttt{\textcolor{blue}{usafarifa97@gmail.com}}; \\\texttt{\textcolor{blue}{risan.aust@gmail.com}}
\\ 
}

\begin{document}

\maketitle
\abstract{
Social media platforms have a vital role in the modern world, serving as conduits for communication, the exchange of ideas, and the establishment of networks. However, the misuse of these platforms through toxic comments, which can range from offensive remarks to hate speech, is a concerning issue. This study focuses on identifying toxic comments in the Bengali language targeting three specific groups: transgender people, indigenous people, and migrant people, from multiple social media sources. The study delves into the intricate process of identifying and categorizing toxic language while considering the varying degrees of toxicity—high, medium, and low. The methodology involves creating a dataset, manual annotation, and employing pre-trained transformer models like Bangla-BERT, bangla-bert-base, distil-BERT and Bert-base-multilingual-cased for classification. Diverse assessment metrics such as accuracy,  recall, precision, and F1-score are employed to evaluate the model's effectiveness. The experimental findings reveal that Bangla-BERT surpasses alternative models, achieving an F1-score of 0.8903. This research exposes the complexity of toxicity in Bangla social media dialogues, revealing its differing impacts on diverse demographic groups.}


\section{Introduction}

Social media sites like LinkedIn, Facebook, Twitter, Instagram, TikTok, and others play a significant role in today's society. It allows individuals to communicate and exchange concepts within a secure environment. Additionally, it functions as a platform for staying updated on trends and ongoing events, as well as for promoting businesses, groups, and social causes. Social media also helps us make friends and talk to people from all over the world, building strong connections\footnote{\url{www.linkedin.com/pulse/importance-social-media-todays-world-johan-smith}}.

Unfortunately, some individuals also misuse these platforms by engaging in toxic comments and harmful behavior. Social media toxic comments refer to harmful, offensive, or negative remarks made by users on social media platforms. These comments can range from personal attacks and insults to hate speech, harassment, and cyberbullying \cite{b2}.  

Toxic language, which is defined as unpleasant, disrespectful, or inappropriate speech that is likely to cause someone to quit a conversation, is a widespread issue on the internet. Using artificial intelligence (ML) models to detect hazardous language in online chats is a hot topic. Models aimed at identifying such online toxicity in chats, however, can be biased \cite{b5}. Recent research has found that several of these models' classification systems are more likely to identify benign language from minority cultures as harmful than identical language representing non-minority groups. Researchers discovered that a publicly available method for detecting toxicity in text has an advantage when predicting high toxicity scores for comments that use African American English when compared to other comments \cite{b3}.
Measuring toxicity among specific identity groups is an essential step in recognizing and addressing online discrimination and harm. By analyzing the type and frequency of toxic comments directed at different identity groups, such as race, ethnicity, gender, or sexual orientation, we gain a deeper understanding of the challenges they encounter in digital spaces.Identifying patterns of harm allows us to assess the severity of the issue and tailor interventions accordingly. For instance, if a particular group consistently faces higher levels of toxic comments, it signals the need for specific support and measures to create a safer environment for them \cite{b6}.
\newline
There has been no such work conducted in the Bengali language regarding toxic comments. The proposed idea is to identify toxic comments used in opposed to a total of three separate groups of people—transgender people, indigenous people and migrated people in Bengali language—from multiple social media.
We may summarize the main points of our study as follows:
\begin{itemize}
\item  Creating a multi-level Bangla  toxic comments Datasets. 
\item  Identifying toxic comments used against three distinct groups of people—transgender people, indigenous people and migrated people.
\item  Measuring the level of toxicity of a comments (high, low and medium). 
\item  Recognizing the subjectivity of toxic comments, where what's harmless to one can be harmful to others.
\end{itemize} 
\section{RELATED WORKS}
Saha et al. \cite{b14} identify  abusive comments written in Bengali on social media platforms. The authors explore various machine learning algorithms and employ SMOTE to address the imbalanced dataset. The study highlights the significance of tackling abusive comments in Bengali and provides valuable insights into effective detection methods.
Haque et al. \cite{b15}  Utilize a method capable of identifying various labels to categorize harmful comments written in the Bangla language. This technique addresses the significance of automating the identification of detrimental content in Bangla and presents an efficient strategy for both classifying and evaluating the intensity of toxic remarks. Jubaer et al.\cite{b11} build a Machine Learning and Deep Learning Approach approche to Classify bangla Toxic comments. Their dataset consists of a total of 250,283 Facebook comments that have been categorized into six types: toxic, insult, obscene, clean and identity hate. Through stemming, the dataset's accuracy may be increased, which might significantly change its accuracy. Rasid et al.\cite{b10} constructs A carefully selected dataset of Bangla offensive language sourced from comments on Facebook referred to as ToxLex\_bn. ToxLex\_bn is a comprehensive wordlist Bigram dataset of Bangladeshi toxic language that has been created from 2207590 Facebook comments. The dataset consists of 8 categories, such as misogynist bullies, sexist, patriarchic, vulgar, political, communal, and racial hate words, each tagged with a binary label (Yes/No) for toxicity. Belal et al.\cite{b9} conduct a pipeline based on deep learning for classifying offensive Bengali comments. The study utilizes a binary classification model using LSTM with BERT Embedding to identify toxic comments, achieving an accuracy of 89.42\%. For multi-label classification, the utilization of a blend between CNN and BiLSTM, coupled with an attention mechanism, results in a performance of 78.92\% accuracy and a F1-score of 0.86. The research also introduces the LIME framework for model interpretability and provides a publicly accessible dataset for further research. Goyal et al. \cite{b6} explore how the self-described identities of human raters affect the detection of toxic language in online comments. 
Groups of raters were created corresponding to identities such as African American and LGBTQ. They found that the identity of the raters greatly impacts how they identify toxicity related to specific identity groups.

\section{Corpus Creation}
\subsection{Data Collection}
The dataset consists of 3100 labeled samples, categorizing individuals into four groups: transgender people, indigenous people, migrated people, and universal toxic. Each category contains comments ranked by toxicity level (high, medium, or low). Comments were manually extracted from TikTok, Facebook, and Instagram, ensuring a diverse range of sources for comprehensive analysis of toxic behaviors and attitudes within the specified groups.

\textbf{Transgender-} The approach for collecting data on toxic comments from the transgender group involved sourcing social media posts from various influencers, TikTok videos, and comments on reels. The data collection focused on identifying comments with toxic behavior, including those that incite violence or express aggression. Additionally, aggressive replies to such comments were also analyzed to ensure a comprehensive dataset for toxic comment classification.

\textbf{Indigenous-} For collecting toxic comment data related to indigenous people, posts from indigenous food vloggers, travel vloggers, and cultural posts were targeted. The focus was on identifying comments that contain toxic behavior, such as aggression or hate speech. Aggressive replies to toxic comments were also considered to provide a more thorough dataset for classifying toxic comments.

\textbf{Migrated-} Toxic comments from the migrated group were sourced from Facebook pages of TV news portals and comments on YouTube videos of news media. The data collection aimed to identify and classify comments with toxic language or aggressive content, including those that wish harm to others or use offensive language.

\textbf{Universal toxic-} To gather data on universal toxic comments, we identified social media posts from different groups and influencer communities. The focus was on finding posts and comments that showed toxic behavior, like inciting violence or expressing aggression. We also looked at aggressive replies to these posts and comments to create a complete dataset for classifying universal toxic comments.

\subsection{Data Annotation}
Data annotation can be done manually by human annotators or using automated tools and techniques. Manual annotation often requires expertise and can be time-consuming, especially for large data sets. Automated annotation methods, such as weak supervision or active learning, can help speed up the process but may require additional validation steps to ensure accuracy \cite{b20}. 

\subsection{\textbf{Identity of annotators}}
The identity of annotators is crucial, as their unique perspectives can impact annotations \cite{b16}. To reduce bias, we selected annotators from diverse racial, geographic, and religious backgrounds. Four annotators—two undergraduate students, one graduate student, and one expert—perform the manual annotations. All are fluent in Bengali, their native language. Table \ref{tab:table2} details their background, area of specialization and other pertinent demographic data. The following essential traits define the annotators:
\begin{itemize}
\item  They are between the ages of 23 and 26. 
\item  None of these individuals are connected to migrated and indigenous groups.

\item  Their experience ranges from one to three years, and their area of study is Natural Language Processing (NLP) and computer vision (CV).

\item They frequently use social media and have seen instances of hostility there.

\end{itemize}
\begin{table*}
\caption{Indentity of the annotators.}
\centering
\begin{tabular}{lllll}
\hline
\textbf{}&
       {\textbf{ANN-1}}&
       {\textbf{ANN-2}}&
       {\textbf{ANN-3}}&
       {\textbf{Expert}}\\
\hline
\textbf{Research-Status} & Undergrad  & Undergrad & Undergrad & NLP researcher \\
\textbf{Research-field} & NLP  & NLP & NLP & NLP, CV, HCI \\
 \textbf{Age} & 23  & 22 & 26 & 35 \\
 \textbf{Gender} & Male  & Female & Male & Male \\
 \textbf{Viewed OnAb} & Yes  & Yes & Yes & Yes \\
  \hline
\end{tabular}

\label{tab:table2}
\end{table*}
Table \ref{tab:table2} is an overview of the annotator's details includes their research status, field of research and personal encounters with online abuse on social media. Here ANN, OnAg denotes annotator and online abuse respectively.

\subsection{\textbf{Annotation guidelines}}
\begin{figure*}[htbp]
\centerline{\includegraphics[scale=0.45]{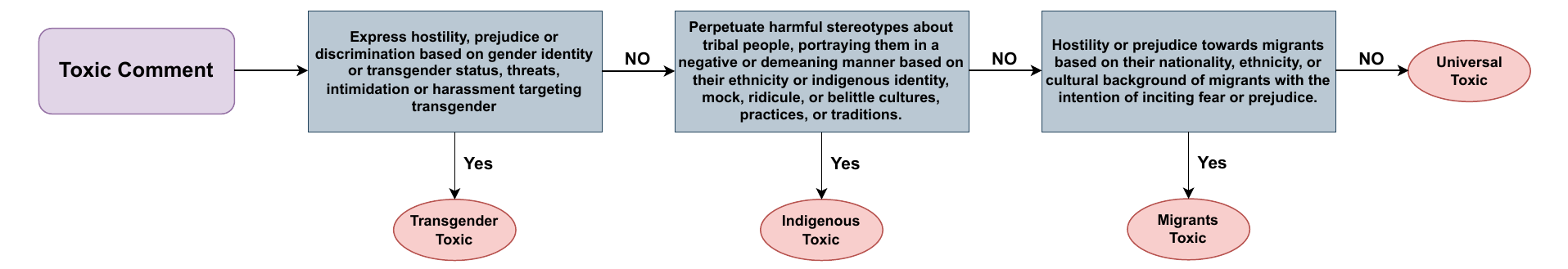}}
\caption{Flow Chart of Annotation guidelines}
\label{Fig:3}
\end{figure*}

Data annotation is critical for training machine learning models because it provides the ground truth information necessary for algorithms to learn. Accurate and consistent annotations are required for high-quality model performance. When annotators are given the opportunity to include their own perspectives, it can provide benefits but also pose issues.\cite{b21}. Creating annotations for a toxic language collection that include sensitive themes like as remarks regarding transgender, tribal, and migratory persons necessitates rigorous ethical thought. The idea is to discover and categorize harmful language into separate groups and degrees. Set specific criteria for poisonous comments for each category, such as abusive language, insulting remarks, hate speech, or discriminatory material. Annotators should evaluate toxicity levels and classify comments as Low, Medium, or High. For example, slang insults or physical bullying may be Medium, but family assaults or death wishes may be High. Comments that do not fit the medium or high criterion should be classed as low toxicity. The Flow Charts of Annotation guidelines is shown \cref{Fig:3}

\begin{enumerate}

\item \textbf{Toxic Comments - Transgender}\newline
Comments that express hostility, prejudice, or discrimination based on a person's gender identity or transgender status.Also a Comments that contain threats, intimidation, or harassment targeting transgender individuals.  
\begin{itemize}
\item Low: {\bng iHjraguela echel na emey ebajha Jay na} (It's not possible to determine the transgender whether they are boys or girls) \newline 
Remarks: Discrimination on gender identity (class)

\item Medium: {\bng E{I} iHjra, dudh na{I} bLaUj perh ik krib}  (He doesn't have any boobs so why is he wearing bra)\newline 
Remarks : Harassment (class) and physical abuse (toxicity level)

\item High: {\bng iHjraguelaek emer ephla Uict}  (The transgenders should be killed)\newline
Remarks: Threat (class) and wish for death(toxicity level)

\end{itemize}

\item \textbf{Toxic Comments - Migrants}\newline
Comments that express hostility or prejudice towards migrants based on their nationality, ethnicity, or cultural background.Also Comments that spread false or harmful information about migrants with the intention of inciting fear or prejudice.
\begin{itemize}
\item Low: {\bng Eedr edsh tYag kra Uict}  (They should leave the country)\newline 
Remarks: identity of nationality (class)
\item Medium: {\bng eraiHNG/ga gulan Aar ktidn Aamaedr edesh thakeb . Eem/net{I} edesh thakar jayga na{I} tarpr Upr E{I} enaNNGra nr/dmar ikTgula Aa{I}sa edesh Dhukes thaket.}  (How long will the Rohingya people stay in our country? There is no place to stay in the country yet these unwanted and disgusting insects like the Rohingya keep entering and living in the country)\newline 
Remarks: identity of nationality (class) and abuse (Toxicity level)

\item High: {\bng eraiHNNGga mruk Era ghnghn bac/ca jnM edy}  (Rohingyas should die because they often give birth to babies)\newline 
Remarks : hostility (class) and death threat (toxicity level)

\end{itemize}

\item \textbf{Toxic Comments - Indigenous people}\newline
Comments that perpetuate harmful stereotypes about tribal people, portraying them in a negative or demeaning manner based on their ethnicity or indigenous identity. Also Comments that mock, ridicule, or belittle tribal cultures, practices, or traditions.

\begin{itemize}

\item Low: {\bng Era sap khay ikbhaeb ich{h}!}  (How do they eat snakes? Shame!)\newline 
Remarks: cultural abuse (class)
\item Medium: {\bng Aaidbaasii gulaer edkhel{I} men cay dhumaya la{I}t/tha{I}. shala banerr bac/ca gula!}  (When I see the indigenous people, I feel like simply beating them. Piece of monkeys!) \newline 
Remarks: mocking (class) and slang (medium)

\item High: {\bng cakma madarecadgula ekaTa ecaday, mruk Era}  (These 'Chakma' motherfuckers only ask for quota, they should die) \newline 
Remarks: demeaning manner (class) and slang,death threat (toxicity level)

\end{itemize}
\item\textbf{Toxic Comments - Universal Toxic}\newline
Toxic comments that are not focused on a particular group like (Tribal People, Transgender, Migrated People) but still exhibit offensive, harmful or disrespectful language. Comments that contain threats, intimidation or harassment towards individuals without focusing on their tribal, transgender or migrant identity. Those comments may not be specifically targeted at tribal, transgender or migrant communities, they can still cause harm and perpetuate harmful behaviors.

\begin{itemize}
\item  {\bng etaedr mt sba{I} Habla paTha na, etaedr ik ekan kaj na{I}, teb elaekr bal kamaga ikchu pysa paib} (Not everyone is lazy like you, don't you have any work, go earn some money)

\item  {\bng nais/tk,,, {O}ek juta marun} (Atheist... beat him with a shoe)

\item  {\bng  kaela emey, edekh{I} eta bim Aaes} (Such a Dark-skinned girl, It makes me vomit)

\end{itemize}

\end{enumerate}
\begin{figure}[h]
\centerline{\includegraphics[scale=0.25]{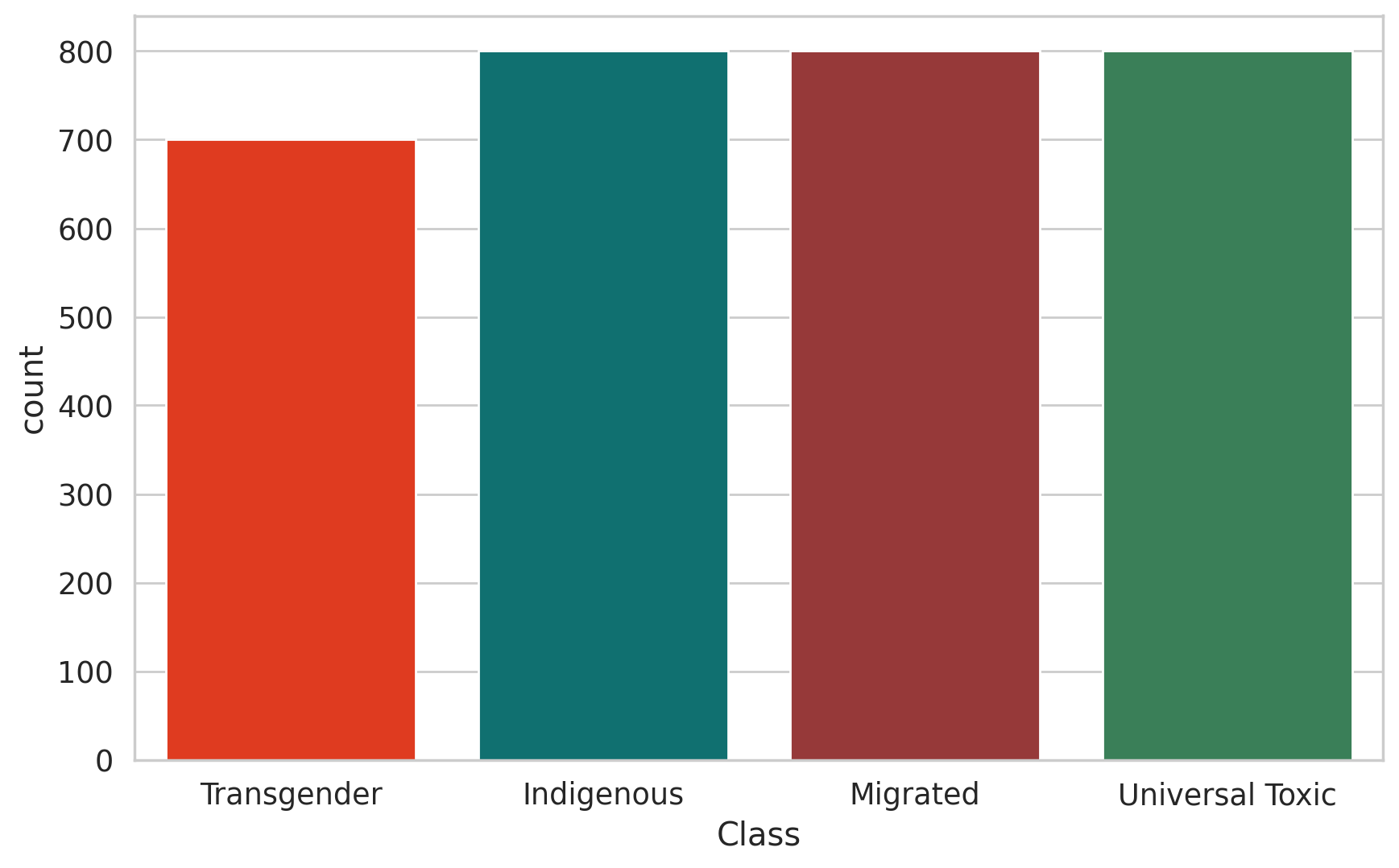}}
\caption{Number of total hate comments by each class}
\label{Fig:2}
\end{figure}

\subsection{Dataset Statistics}
The collection of data comprises a complete sum of 3100 data instances. Among these instances,  2300 are labeled as Toxic comments, while the remaining 800 are labeled as universal toxic comments. These Toxic texts are further categorized into 3 classes, where the Trans, indigenous , and Migrants classes have 700, 800, and 800 text samples, respectively. The dataset overview is summarized in \Cref{Fig:2}. To request access to the dataset, please reach out to the corresponding author.


\section{PROPOSED METHODOLOGY}
\begin{figure}[h]
    \includegraphics[width=1\textwidth ]{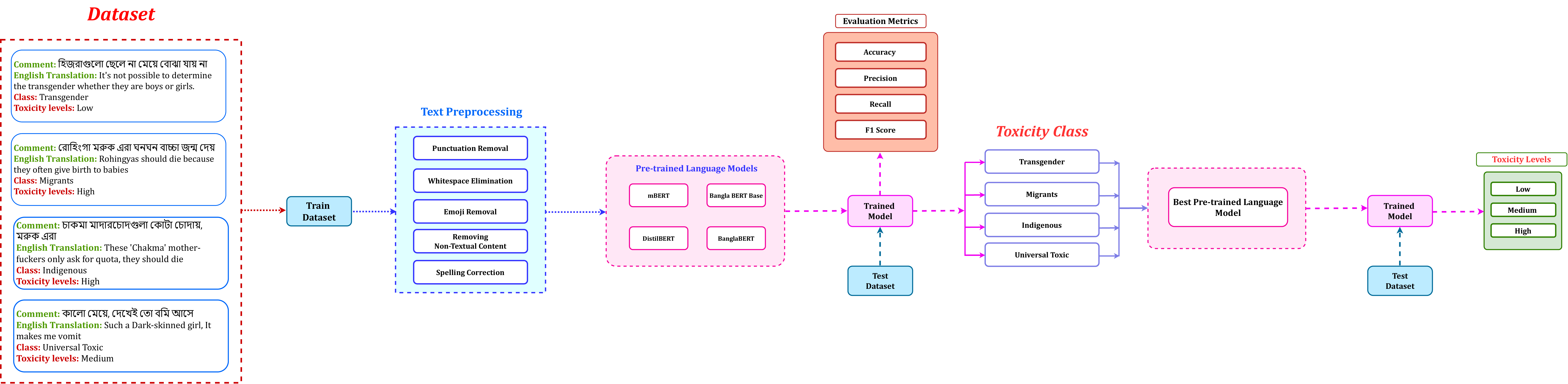}
    \caption{The diagram showcases the suggested methodology for Unveiling the Levels of Toxicity in Bangla Social Media Comments}
    \label{Fig:1}
\end{figure}
In this section, we outline a suggested methodology for Unveiling the Levels of Toxicity in Bangla Social Media Comments. The schematic representation of the proposed method for identifying bangla toxic comment in \Cref{Fig:1}.

To effectively handle toxic comment detection in Bangla, The dataset is preprocessed to normalize Bengali text by managing whitespace, commas, and removing extraneous letters, with a concentration on Bangla Unicode characters. During fine-tuning, pre-trained language models such as mBERT \cite{mbert}, DistilBERT \cite{Distil}, bangla-bert-base \cite{banglabertbase}, and BanglaBert \cite{BanglaBert} are trained on a Toxic Bangla dataset using transfer learning techniques, and model parameters are adjusted using gradient descent optimization using the AdamW optimizer and CrossEntropyLoss. Hyperparameters like as learning rate of 2e-5, batch size of 16 are tuned to improve model performance while avoiding overfitting. The model's performance is measured using accuracy \cite{acc}, precision \cite{pre}, recall \cite{recall}, and F1-score \cite{f1} measures.

\section{ EXPERIMENTAL RESULTS} 
To classify toxic comments targeting specific groups such as transgender, Indigenous, migrants and universal toxic, we utilized several pre-trained language models. Here in table \ref{tab:table3} BanglaBERT performed the best, achieving an accuracy of 0.8903. The other models, DistilBERT base multilingual, Bangla BERT base and BERT base multilingual, obtained accuracies of 0.8323, 0.8645 and 0.8419 respectively. In table \ref{tab:table4}, we presented the classification of toxicity levels for each category using the best model, BanglaBERT. The accuracy for transgender across low, medium and high toxicity level is 0.6714. For the indigenous, the accuracy stands at 0.6375 while for the Migrants, it is 0.5625. These results highlight BanglaBERT's superior performance, which can be attributed to its ability to capture linguistic nuances in Bangla, as well as its training on a larger and more diverse Bangla corpus compared to the other models.
\begin{table}[ht]
\begin{center}
 \caption{ Performance comparison of Pre-trained Language Models For Bangla Toxic Comment Classifications }
  \label{tab:table3}
  \renewcommand{\arraystretch}{1.5}
   
  \begin{tabular}{@{}c@{\hspace{10pt}}c@{\hspace{10pt}}c@{\hspace{10pt}}c@{\hspace{10pt}}c@{}}
   
     \hline
       \textbf{Models}&
       \textbf{\textbf{Accuracy}}&
       \textbf{\textbf{Precision}}&
       \textbf{\textbf{Recall}}&
       \textbf{\textbf{F1 Score}}\\

        \hline   
         DistilBERT & 0.8323    &  0.8370    & 0.8320  &  0.8320    \\
        \hline 
         bangla-bert-base & 0.8645  & 0.8651 & 0.8696 & 0.8648\\
        \hline
          \textbf{BanglaBERT} &  \textbf{0.8903} & \textbf{0.8906}  &  \textbf{0.8901}  &  \textbf{0.8903} \\
        \hline 
        mBERT &  0.8419 & 0.8433  &  0.8478  &  0.8419 \\
        \hline

    \end{tabular}
    \end{center}
 
\end{table} 

\begin{table}[h]
\begin{center}
 \caption{ Evaluation Metrics for Classifying Toxicity Levels in Bangla Social Media Comments }
  \label{tab:table4}
  \renewcommand{\arraystretch}{1.5}
   
  \begin{tabular}{c@{\hspace{5pt}}cc@{\hspace{5pt}}c@{\hspace{5pt}}c@{\hspace{5pt}}c@{\hspace{5pt}}c}
   
     \hline
       \textbf{Class}&
        \textbf{Models}&
       \textbf{\textbf{Accuracy}}&
       \textbf{Level}&
       \textbf{\textbf{Precision}}&
       \textbf{\textbf{Recall}}&
       \textbf{\textbf{F1 Score}}\\

        \hline   
         & &    & Low   &  0.7143    & 0.7143  &  0.7143    \\
      Transgender & BanglaBERT & 0.6714 & Medium     &  0.5769     & 0.6522  & 0.6122   \\
         &  &   & High    &  0.7778   & 0.5833  &  0.6667    \\
       
        \hline 
        &  &   & Low   &  0.7778    & 0.5526  &  0.6462   \\
    Indigenous  & BanglaBERT &  0.6375 & Medium     &  0.6286     & 0.7097  &  0.6667    \\
        &  &     & High    &  0.4444    & 0.7273  &  0.5517    \\
       
        \hline 
     & &    & Low   &  0.5745    & 0.6585  &  0.6136    \\
     Migrants    & BanglaBERT & 0.5625 & Medium     & 0.5600     & 0.4516  &  0.5000    \\
         &  &   & High    &  0.5000    & 0.5000  &  0.5000    \\
       
        \hline 
    

    \end{tabular}
    \end{center}
 
\end{table} 

\begin{figure}[h]
    \includegraphics[width=1\textwidth ]{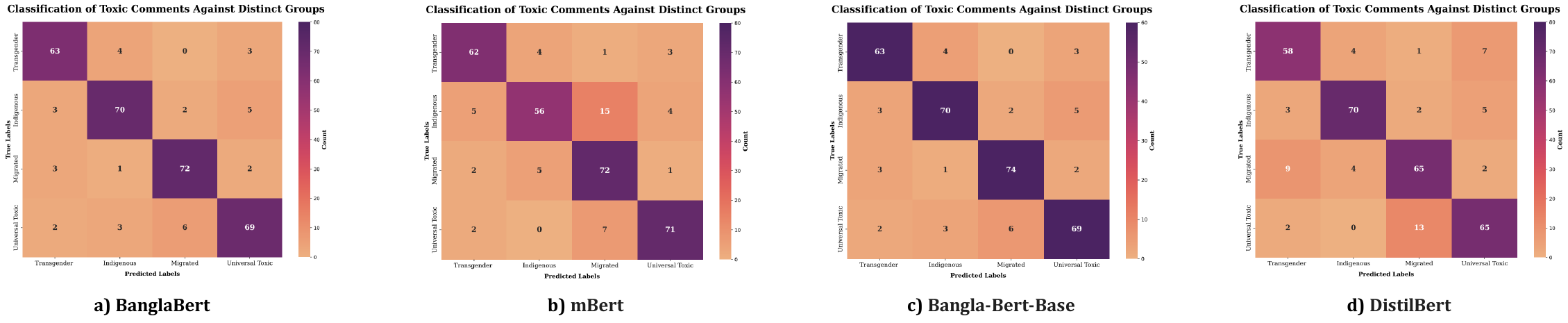}
    \caption{Confusion matrix of Pre-trained Language Models For Bangla Toxic
Comment Classifications}
    \label{fig:error}
\end{figure}

\section{Conclusion}
To sum up, this paper emphasizes addressing toxic comments in a multicultural context, targeting groups like transgender individuals, indigenous people, and migrants. While previous research has concentrated on toxicity detection in languages such as English and others, this study makes a unique contribution to the underexplored field of Bangla language toxicity detection, particularly in vulnerable groups. The toxicity identification of languages like English and other languages has been the subject of previous research, but this study makes a new contribution to the underexplored field of Bangla toxicity detection especially in vulnerable groups. The study presents a manually annotated multi-level dataset and evaluates several transformer-based models, including Bangla-BERT, bangla-bert-base, mBert and distilBert. Bangla-BERT showed the highest precision, accuracy, recall, and F1-score. Future plans include expanding the dataset to include more groups and refining toxicity levels (very high, high, low, very low) for a more nuanced understanding of toxic comments.

\end{document}